\documentclass{article}
\usepackage{spconf,amsmath,epsfig,graphicx}
\usepackage{xcolor}
\usepackage[autopunct=true]{csquotes}

\newcommand{\tracy}[1]{\textcolor{black}{#1}}

\newcommand{\daniel}[1]{\textcolor{black}{#1}}
\newcommand{\henry}[1]{\textcolor{black}{#1}}

\let\OLDthebibliography\thebibliography
\renewcommand\thebibliography[1]{
  \OLDthebibliography{#1}
  \setlength{\parskip}{0pt}
  \setlength{\itemsep}{0pt plus 0.3ex}
}

\begin{document}\sloppy

\def\x{{\mathbf x}}
\def\L{{\cal L}}

\title{Attention-Aware Anime Line Drawing Colorization}
%
\name{Yu Cao, Hao Tian, P.Y. Mok$^{\ast}$\thanks{$^{\ast}$Corresponding author}}
\address{The Hong Kong Polytechnic University,\\ \{yu-daniel.cao, hao-henry.tian\}@connect.polyu.hk, tracy.mok@polyu.edu.hk}

\maketitle

\begin{abstract}
Automatic colorization of anime line drawing has attracted much attention in recent years since it can substantially benefit the animation industry. User-hint based methods are the mainstream approach for line drawing colorization, while reference-based methods offer a more intuitive approach. Nevertheless, although reference-based methods can improve feature aggregation of the reference image and the line drawing, the colorization results are not compelling in terms of color consistency or semantic correspondence. In this paper, we introduce an attention-based model for anime line drawing colorization, in which a channel-wise and spatial-wise Convolutional Attention module is used to improve the ability of the encoder for feature extraction and key area perception, and a Stop-Gradient Attention module with cross-attention and self-attention is used to tackle the cross-domain long-range dependency problem. Extensive experiments show that our method outperforms other SOTA methods, with more accurate line structure and semantic color information.
\end{abstract}

\section{Introduction}
\label{sec:intro}
Image colorization has attracted extensive research attention~\cite{XiaHWW22, huang2022unicolor, lu2020gray2colornet} in the field of Computer Graphics and Multimedia. Line drawing colorization is an essential process in the animation industry. For example, manga and cartoon line drawing colorization is a practical application of this, while manual colorizing is time consuming, especially for line drawings with complex structure. In the past, different methods~\henry{\cite{qu2006manga, varga2017automatic,sykora2009lazybrush}} have been proposed to speed up the process. Line drawing colorization is challenging, because line drawings, different from grayscale images, only contain structure content composing of a series of lines without any luminance or texture information.

\begin{figure}[t] 
\centering 
\includegraphics[width=0.48\textwidth]{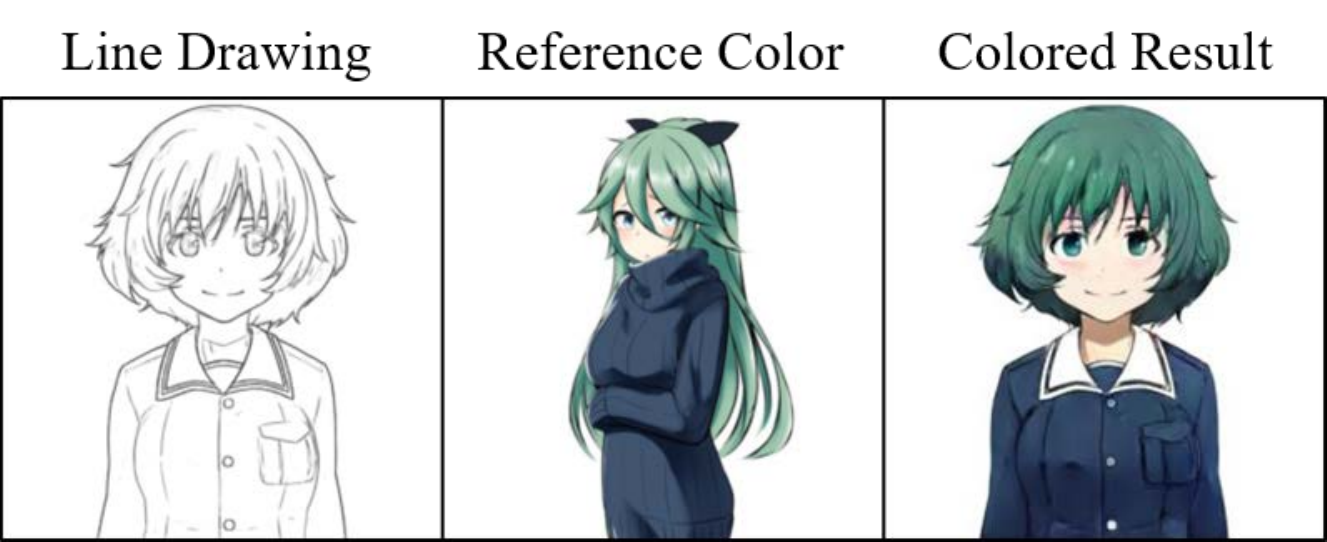} 
\caption{Sample images of line drawing, reference color image, and generated colored result of our model are from left to right. All three images have the resolution of $256\times256$.} 
\label{Fig.1} 
\end{figure}

The colorizing process can be generally regarded as a conditional image-to-image translation problem that directly converts the input line drawings to the output colored results. Early work \cite{varga2017automatic} utilized a neural network to automatically colorize the cartoon images with random colors, which is the first deep learning based cartoon colorization method. Nevertheless, many interactions are usually needed to refine the colored results to satisfy what the user specified. To effectively control the color of the result, many user-hint based methods have been proposed successively, such as point colors \henry{\cite{adeleine, ci2018user, yuan2021line}}, scribble colors \cite{petalica}, text-hint \cite{kim2019tag2pix}, and language-based \cite{zou2019language}. These user-hint based methods are still not convenient or intuitive, especially for amateur users without aesthetic training.

Reference-based colorization methods~\cite{chen2020active, lee2020reference, li2022eliminating} provide a more convenient way, which can automatically complete the colorizing process without other manual intervention. Users only need to prepare a line drawing and a corresponding reference color image, see Fig.~\ref{Fig.1} as an illustration. In the typical approach, two encoders are used to extract the line drawing feature and reference color feature respectively, and then a feature aggregation block is designed to inject color from the corresponding position of reference image into line drawing. Lee \emph{et al.}~\cite{lee2020reference} proposed an attention-based Spatial Correspondence Feature Transfer (SCFT) module, and quantitatively proved its feature aggregation ability is superior to addition block and AdaIN~\cite{huang2017arbitrary} block. Li \emph{et al.}~\cite{li2022eliminating} eliminated the gradient conflict among attention branches using Stop-Gradient Attention (SGA) module.
Meanwhile, these methods bring new challenges in the color consistency and semantic correspondence between the colored image and reference image. It is attractive to design a line drawing colorization algorithm that meets the above two conditions, which will greatly reduce the tedious work of animators.

In this paper, we propose a novel conditional adversarial colorization network combining Convolutional Attention module and Stop-Gradient Attention module trained fully on anime line drawing dataset~\cite{anime}. \daniel{We first design a line drawing extraction network based on this dataset which can be used to augment more data.} Then, we train our colorization model through a proxy task of line drawing guided distorted image restoration. The encoder equipped with Convolutional Attention module is introduced to help the encoder extract exact multi-scale features. Therefore, the colorization results of our model have both geometry details and accurate colors. Meanwhile, Stop-Gradient Attention module shows great feature aggregation performance and gives our model better semantic correspondence ability. With the attention-based network, we can generate colored results which are comparable to manual coloring results (see an example in Fig.~\ref{Fig.1}). Our main contributions can be summarized as follows:
\begin{itemize}
    \item We \daniel{design an U-Net~\cite{ronneberger2015u} based network} to extract line drawing from color images to simulate manually drawn ones. Since the number of available anime line drawing paired dataset is limited, \daniel{we can use this model to extend the number of current data.}
    \item We \tracy{develop an} anime line drawing colorization method \tracy{based on attention-aware mechanism by integrating} Convolutional Attention module and Stop-Gradient Attention module \tracy{so as to improve the color consistency and semantic correspondence of coloring results}. 
    \item \tracy{Both qualitative and quantitative comparative analyses show that} our method has outperformed other reference-based state-of-the-art methods in the task of anime line drawing colorization via training on a proxy task of line drawing guided distorted image restoration.
\end{itemize}

\section{Related Work}
\subsection{Line Drawing Colorization}
Since line drawing contains only structure information with sparse line sets, existing colorization methods for gray-scale images cannot be used directly. Traditional line drawing colorization approaches~\cite{qu2006manga, sykora2009lazybrush} are commonly optimized-based, allowing users to use brushes to inject desired colors into specific regions. With the advancement of deep learning and for better control of color, many user-hint colorization methods \cite{ci2018user, kim2019tag2pix, zou2019language} spring up. However, the complexity of such user-hint methods will become more labor-intensive as the number of line drawings increases, and many interactions are also needed. Therefore, many reference-based colorization methods~\cite{chen2020active, lee2020reference, li2022eliminating, furusawa2017comicolorization} have been proposed, which are very suitable for colorizing line drawing sets or videos of anime characters. Our proposed colorization model belongs to this reference-based method, and we can generate more faithful results with fantastic visual quality.

\subsection{Semantic Correspondence}
Semantic correspondence~\cite{xiao2022learning} is one of the fundamental problems in computer vision whose goal is to establish dense correspondences across images containing the targets of the same category or with similar semantic information. Reference-based line art colorization can also be viewed as cross-domain correspondence since the texture difference between line drawing and reference color image. 
Most semantic correspondence learning methods are designed for the grayscale image colorization. However, our proposed method relies on the reasonable design of the attention mechanism module, which can achieve plausible results for anime line drawing colorization. 

\subsection{Attention Mechanism}
With the attention mechanism~\cite{hu2018squeeze, hu2021a2} showing good performance in feature extraction and aggregation in image perception task, it makes neural model much closer to the human visual perception system. Recently some attention-based line drawing colorization methods~\cite{yuan2021line, lee2020reference, li2022eliminating} have been proposed for improving the model’s colorization ability. 
For grayscale image colorization, Kumar~\emph{et al.}~\cite{kumar2021colorization} presented the Colorization Transformer that entirely relies on self-attention for image colorization. 
Since few research work focuses on line drawing colorization using attention-based method, we mainly compare our method with Lee~\emph{et al.}~\cite{lee2020reference} and Li~\emph{et al.}~\cite{li2022eliminating}, and our method achieves the state-of-the-art result.

\begin{figure*}[htbp] 
\centering 
\includegraphics[width=0.95\textwidth]{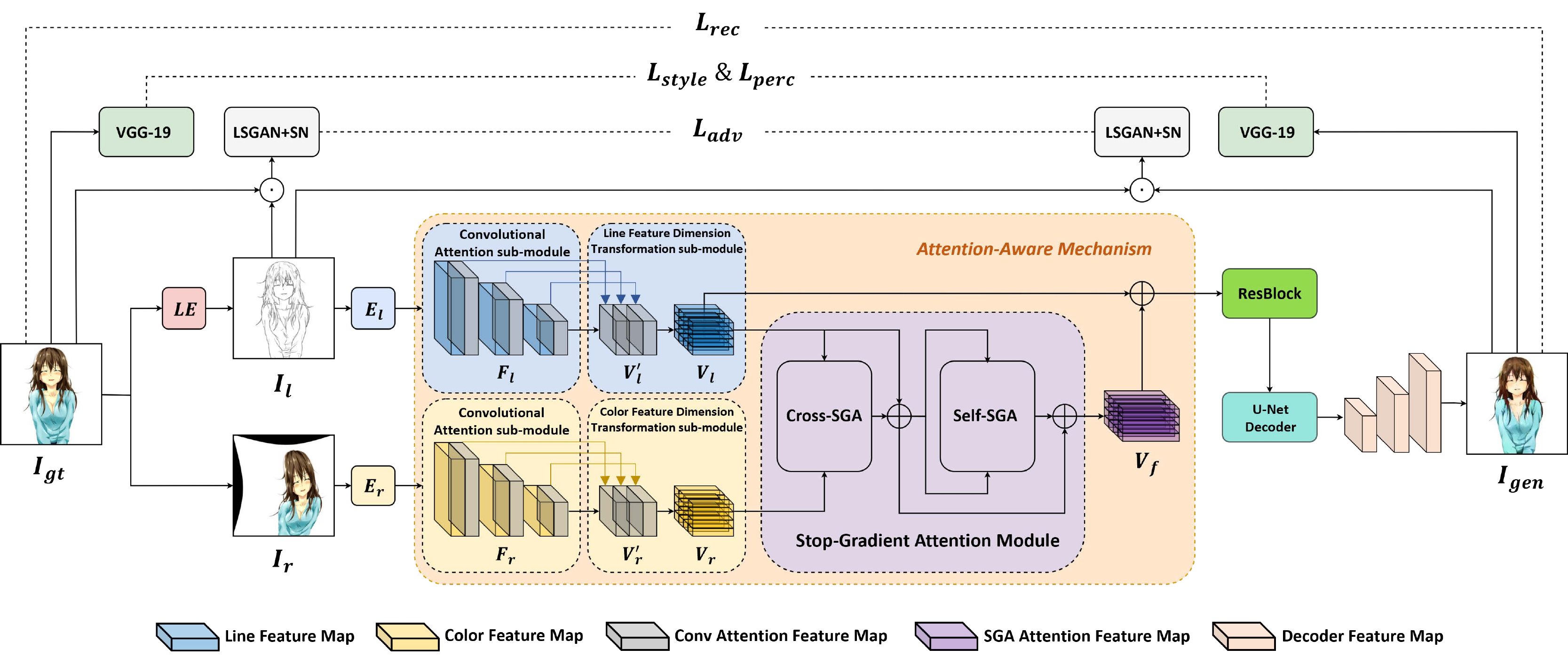}   
\caption{\daniel{The overview training pipeline of our network.}} 
\label{Fig.2} 
\end{figure*}

\section{Method}
Given a line drawing and a reference color image, our network can generate a colored result which contains clear geometry structure of the line drawing and accurate semantic color of the reference image. We formulate the problem as a novel conditional adversarial network in which the training process itself is \tracy{regarded} as line drawing guided distorted reference color image restoration, as shown in Fig.~\ref{Fig.2}.
Inspired by~\cite{lee2020reference}, we train our model in a self-augmented supervised manner, enabling the network to be trained on limited paired data. The trained model can perform reference-based colorization for other line drawing inputs during inference. 

\subsection{Model Architecture}
As illustrated in Fig.~\ref{Fig.2}, assuming $I_{gt}$ is an original colored image, and an line drawing $I_l$ is extracted from $I_{gt}$ using line drawing extraction network $LE$. It is important to note that there are usually large spatial structure discrepancy between the reference color image and the line drawing, thus a large volume of paired data of before and after colorization is needed to train a colorization network. Nevertheless, there are limited data of amine line drawing and it is very costly to prepare a dataset with such paired data. To address this problem, we first train a LE network to extract line drawings from colored amine images using an existing dataset~\cite{anime}. With trained LE network, we augment data by generating corresponding anime line drawings from color images obtained from the internet.

Our model consists of a generator and a discriminator \tracy{as well as a specially designed \textquote{Attention-aware mechanism}}. For the generator, there are two encoders $E_l$ and $E_r$, which are used to extract feature from $I_l$ and $I_r$, respectively. Note that the input channel number of $E_l$ is 1, while the input channel number of $E_r$ is 3. By designing two feature extractors, the line feature and color feature of the image are disentangled~\cite{huang2018multimodal}. The extracted line drawing feature maps $F_l$ and color style feature maps $F_r$ will go through a Convolutional Attention sub-module and a Feature Dimension Transformation sub-module to get multi-scale attention feature map $V_l$ and $V_r$. Next, we adopt the Stop-Gradient Attention (SGA) module to fuse $V_l$ and $V_r$ to get the fusion feature $V_f$. As shown in Fig.~\ref{Fig.2}, the SGA module contains cross-attention and self-attention blocks to better acquire dense semantic correspondence and to tackle cross-domain long-range dependency problem. Then $V_f$ and $V_l$ will add together and pass through several residual blocks, and followed by a U-Net~\cite{ronneberger2015u} decoder to generate restored result $I_{gen}$. 

For the discriminator, we utilize conditional LSGAN~\cite{mao2017least} combined with Spectral Normalization(SN)~\cite{miyato2018spectral} for stable training. 
Both generator and discriminator use Batch Normalization (BN) to accelerate the training speed, but there is no BN or SN in the final layer of the discriminator.

\subsection{Attention-Aware Mechanism}
\daniel{Our model performs colorization in such as way that the colored result contains semantic color information and style from the reference color image while representing the same visual content as the target line drawing. To this end, our model should have mechanisms to
learn three essential features: 1) the content of the line drawing,
2) the color information of the reference image, and 3) the visual correspondences
between the line drawing and the reference image. We use the attention mechanism to learn these features.}
\subsubsection{Convolutional Attention \tracy{(CA)} Module}
To improve the feature extraction ability of the two encoders $E_l$ and $E_r$, which take into account both global and local features, we add spatial and channel attention mechanisms~\cite{woo2018cbam} behind each convolution layer of the encoder. \daniel{We choose the ~\cite{woo2018cbam} attention module because it is computationally more efficient and requires less learnable parameters.} This operation enables the network to adaptively extract important features from reference image and line drawing, thus the colored results will have accurate colors, natural transitions with clear line structure. After this, the multi-scale features will be integrated in Feature Dimension Transformation sub-module~\cite{woo2018cbam}. More specifically, the features from the Convolutional Attention sub-module will be resized to the size of the final layer feature map, concatenate respectively on the channel dimension, and then transpose their spatial dimension with the channel dimension to generate $V_l$ and $V_r$, 
which contains multi-scale feature.

\subsubsection{Stop-Gradient Attention \tracy{(SGA)} Module}
\daniel{To learn visual correspondences between the colorized image and reference sketch, we build elf-attention based feature aggregation module. 
Since there exists a gradient conflict issue in self-attention based feature aggregation module, we utilize the Stop-Gradient Attention module~\cite{li2022eliminating} to integrate the line drawing feature and reference color image feature.} The key operation is truncating the back propagation of the gradient of attention matrix. Inside the SGA module, the attention matrix is normalized through row dimension and column dimension respectively, and we use two SGA module equipped with short connections for cross-attention and self-attention separately. 
By integrating SGA module with convolutional attention module, our model shows good performance in semantic correspondence and generates colored results in high quality. 

\subsection{Loss Function}
To train our network for high quality colorization results, namely with better color consistency and semantic correspondence, we define the loss function as follows.

\emph{\textbf{L1 Loss.}} Since the main training task of our model is distorted image restoration, the pixel-level reconstruction loss is needed. We use L1 loss rather than L2 loss as L1 encourages less blurring, according to~\cite{isola2017image}. We directly use it to penalize the model for the pixel-level loss between the restored image $I_{gen}$ and original image $I_{gt}$:
\begin{eqnarray}
    L_{rec} = E[\parallel G(I_l , I_r) - I_{gt} \parallel_1]
\end{eqnarray}
where $G$ denotes the generator.

\emph{\textbf{Perceptual Loss}.} We use the VGG-19 model pretrained on ImageNet as feature extractor to compute the high-level semantic and low-level perceptual loss~\cite{johnson2016perceptual} through multiple feature layers. The loss function is defined as follows:
\begin{eqnarray}
    L_{perc} = E[\sum\limits_{l}\parallel \phi_l(I_{gen}) - \phi_l(I_{gt}) \parallel_1]
\end{eqnarray}
where $\phi_l$ represents the $l^{th}$ feature layer after ReLU activation function of VGG-19.

\emph{\textbf{Style Loss}.} To generate image with more clear appearance and to address checkboard artifacts, a style loss~\cite{johnson2016perceptual} is introduced as follows:
\begin{eqnarray}
    L_{style} = E[\sum\limits_{l}\parallel \mathcal{G}(\phi_l(I_{gen})) - \mathcal{G}(\phi_l(I_{gt})) \parallel_1]
\end{eqnarray}
where $\mathcal{G}$ means the gram matrix.

\emph{\textbf{Adversarial Loss}.} Since our model is a kind of generative model, in order to improve the generation ability and make the training process more stable, 
we leverage the conditional LSGAN~\cite{mao2017least} objective function equipped with Spectral Normalization (SN) to calculate an adversairal loss:
\begin{eqnarray}
    L_{adv} = E[\parallel D(I_{gt}, I_l) \parallel_2^2] + E[\parallel 1 - D(G(I_l, I_r), I_l) \parallel_2^2]
\end{eqnarray}
where $D$ denotes the discriminator.

\emph{\textbf{Total loss function.}} We set the optimization goal by combining all the above losses as follows:
\begin{eqnarray}
\begin{split}
\mathop{min}\limits_G\mathop{max}\limits_DL = \lambda_{adv}L_{adv} + \lambda_{rec}L_{rec} + \\ \lambda_{perc}L_{perc} + \lambda_{style}L_{style}
\end{split}
\end{eqnarray}
where \tracy{$\lambda_{adv}$, $\lambda_{rec}$, $\lambda_{perc}$ and $\lambda_{style}$ represent the corresponding weights of the adversarial loss, L1 loss, perceptual loss and style loss. }
We set $\lambda_{adv}=1$, $\lambda_{rec}=30$, $\lambda_{perc}=0.01$, and $\lambda_{style}=50$.

 \begin{figure*}[t]
 \centering  
 \includegraphics[width=1.0\textwidth]{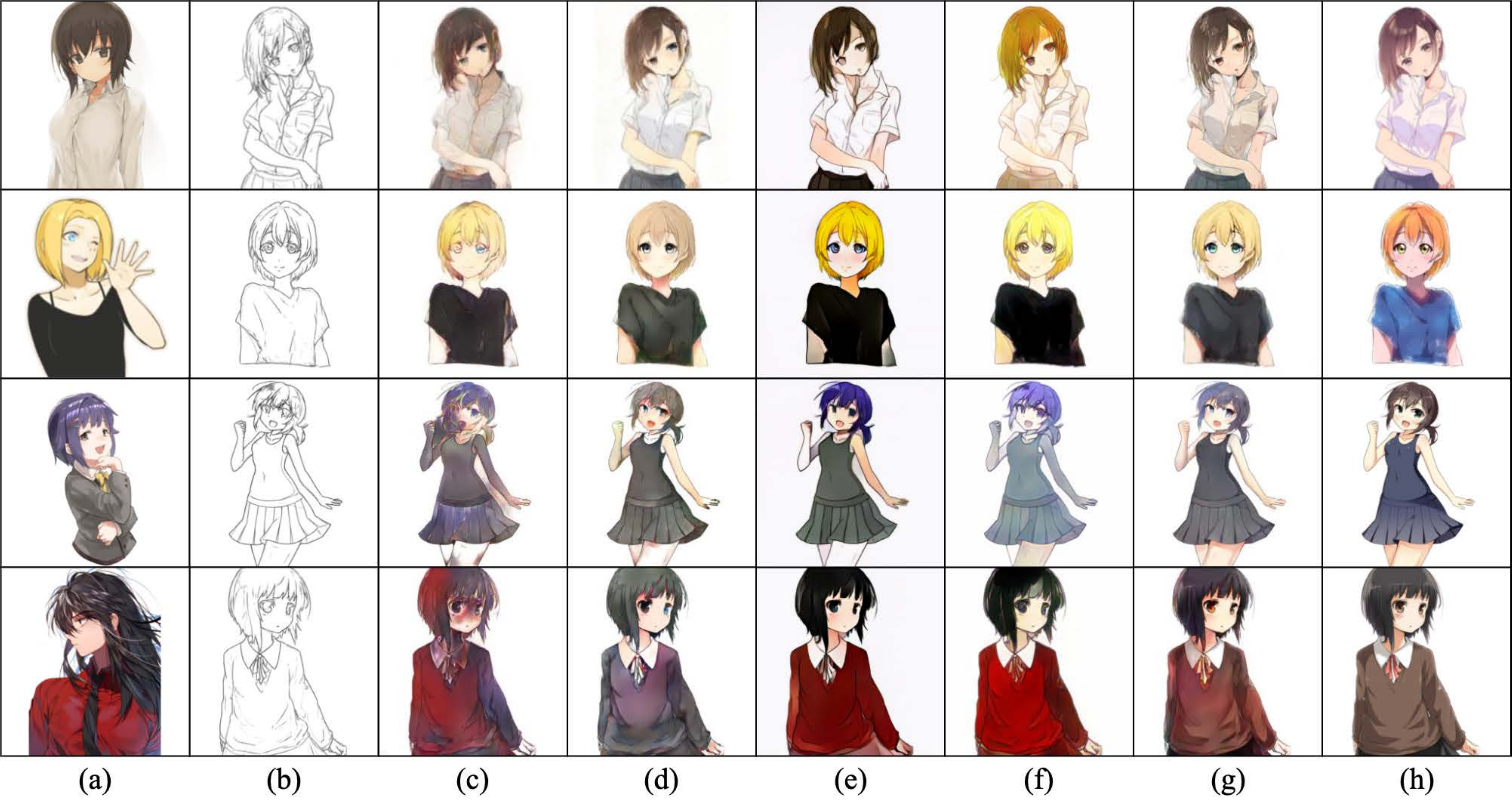} 
\vspace{-15px}
\caption{Qualitative result comparison: (a) reference color images, (b) line drawings, (c) the results of Lee \emph{et al.}~\cite{lee2020reference}, (d) the results of Li \emph{et al.}~\cite{li2022eliminating}, (e) the results of petalica~\cite{petalica}, (f) the results of adeleine~\cite{adeleine} , (g) our results, and (h) original color images}
\label{Fig.3}
 \end{figure*}

\section{Experiments}
\subsection{Data Preparation}
We trained our model with an anime dataset consisting of 17769 images shared on Kaggle website~\cite{anime}. We first trained a line drawing extraction network based on this data which can be used to perform data augmentation. We use TPS transformation \cite{bookstein1989principal} to convert the original color anime image to a geometrically distorted image online along with the training of our colorization network (Fig.~\ref{Fig.2}). The full data are used to train our model without separating validation data. We can use our trained line drawing extraction network to prepare extra data for testing.

\subsection{Implementation Details}
We implemented our method using PyTorch framework and the model was trained on a NVIDIA 3090 GPU with a batch size of 16. We set the total iteration number as 250000, and both the generator and the discriminator were alternately updated in every iteration. The input size of all images was set at $256\times256$ and the pixel value was normalized to the range of [-1, 1].  Weight parameters for five ReLU activated feature layers were all set to 1 for computing the perceptual loss and style loss. We used Adam optimizer with $\beta_1$=0.5, $\beta_2$=0.999. The learning rates of generator and discriminator were set to 0.0001 and 0.0002, respectively, and we did not use extra learning rate update strategy during training.

\subsection{Qualitative Evaluation}
Fig.~\ref{Fig.3} illustrates the visual comparisons between our method and two other state-of-the-art methods trained on the same dataset~\cite{anime} \daniel{and two user-hint colorization tools.}
For a fair comparison, all three models \daniel{(two tools are not learning based methods)} are pretrained to the extent of convergence with batch size of 16. As shown in Fig.~\ref{Fig.3}, Lee \emph{et al.}~\cite{lee2020reference} generates results with color bleeding, especially in the region of hair and clothes.
Li \emph{et al.}~\cite{li2022eliminating} generates some results with distorted color compared with the reference images. 
\daniel{\cite{petalica} and \cite{adeleine} are two user-hint colorization methods using color scribbles and color points as input respectively.} \daniel{They are not very convenient to use because manually specifying the colors from the reference image is imprecise and prone to color aliasing and semantic interference problems during the coloring process.}
\daniel{In contrast, reference-based colorization method has obvious advantages, which can accurately transfer the color information and color style from the reference images to the line drawings.}
Our method offers good semantic correspondence in the parts of hair, eyes and clothes. 
In addition, we also find from the experiments that our model with Convolutional Attention mechanism converges faster than~\cite{li2022eliminating} does during the training stage.

\subsection{Quantitative Evaluation}
We set two kinds of coloring cases to evaluate the performance of our network, self-reference colorization and random-reference colorization. For self-reference colorization, the line drawing and reference image are paired, ideally the colorized output should be exactly the same as the reference image. In order to evaluate the colorization quality, we measure the structure similarity and perceptual similarity between the colored result and reference image using Peak Signal-to-Noise Ratio (PSNR), Multi-Scale Structural Similarity Index Measure (MS-SSIM) and Learned Perceptual Image Patch Similarity (LPIPS)~\cite{zhang2018unreasonable}. For the random-reference colorization, it is more like the common practical usage when using exemplar-based colorization method. To evaluate the generative ability of GAN-based network, we perform a quantitative study by calculating the Fréchet Inception Distance (FID) score between the colorized result and reference image. A smaller FID indicates that the distribution of the colored image is closer to the 
reference color image. 

As is shown in Table~\ref{tab 1}, our method achieves the best FID, 17.35\% improvement in comparison to \cite{li2022eliminating} and 50.55\% improvement in comparison to \cite{lee2020reference}. For three metrics of self-reference colorization, our method is still the best one. It indicates that through training on the proxy task of image restoration, our model not only acquires good image restoration performance but also can be used to line drawing colorization.

\subsection{Ablation Study}
\daniel{Table~\ref{tab 1} also gives ablation study results, in which Lee~\emph{et al.}~\cite{lee2020reference} can be regarded as baseline, while Li~\emph{et al.}~\cite{li2022eliminating} utilizes SGA module only, our method integrates both CA and SGA modules together. It shows that CA module itself (i.e. \cite{lee2020reference}+CA) or SGA module itself (i.e. \cite{li2022eliminating}) are not as effective as the current proposal of attention-aware mechanism that integrates both.}
\begin{table}[h]
\begin{center}
\caption{\tracy{Quantitative comparison with SOTA methods.}}
\vspace{-5px}
\label{tab 1}
\resizebox{\linewidth}{!}{
\begin{tabular}{c|cccc}
   \hline
   Method & FID$\downarrow$ & PSNR$\uparrow$ & MS-SSIM$\uparrow$ & LPIPS$\downarrow$\\
   \hline
   \hline
    Lee \emph{et al.}~\cite{lee2020reference}& 47.20 & 26.36 & 0.95 & 0.06 \\
    Lee \emph{et al.}~\cite{lee2020reference} + CA  & 40.08 & 24.70 & 0.92 & 0.09 \\
    Li \emph{et al.}~\cite{li2022eliminating}  & 28.24 & 21.01 & 0.94 & 0.07 \\
    Ours & \textbf{23.34} & \textbf{27.21} & \textbf{0.96} & \textbf{0.05} \\
   \hline  
\end{tabular}
}
\end{center}
\end{table}

\section{Conclusion}
In this paper, we proposed a new attention-based colorization algorithm for anime line drawings. To do so, we first train a line drawing extractor using existing anime dataset available from the internet. The trained line drawing extractor is integrated in the model training of our proposed attention-based colorization network for data augmentation. Our method achieves faithful colorization results using a novel attention mechanism by integrating (1) a Convolutional Attention module containing channel-wise and spatial-wise attention block, and (2) a Stop Gradient-Attention module including cross-attention and self-attention block. Through extensive experiments, our method has demonstrated better performance both qualitatively and quantitatively, outperforming other state-of-the-art methods, with more accurate line structure and semantic color information.

\section{Acknowledgements}
The work described in this paper was supported by grant from the Research Grants Council of the Hong Kong Special Administrative Region, China (Grant Number 152112/19E); it was also partially support by the Innovation and Technology Commission of Hong Kong under grant ITP/028/21TP.


\renewcommand{\footnotesize}{\fontsize{8}{8.5}\selectfont}
\footnotesize\bibliographystyle{IEEEbib}
\bibliography{reference}

\end{document}